# An efficient genetic algorithm for large-scale planning of robust industrial wireless networks

Xu Gong [a,*], David Plets [a], Emmeric Tanghe [a], Toon De Pessemier [a], Luc Martens [a], Wout Joseph [a]

[a] *Department of Information Technology, Ghent University/imec, Technologiepark 15, 9052 Ghent, Belgium*

\* E-mail*:* xu.gong@ugent.be. Tel.: +32 4 88 69 68 01. Fax.: +32 9 33 14899.

E-mail of co-authors: david.plets@ugent.be, emmeric.tanghe@ugent.be, toon.depessemier@ugent.be, luc.martens@intec.ugent.be, wout.joseph@ugent.be

**Abstract**

An industrial indoor environment is harsh for wireless communications compared to an office environment, because the prevalent metal easily causes shadowing effects and affects the availability of an industrial wireless local area network (IWLAN). On the one hand, it is costly, time-consuming, and ineffective to perform trial-and-error manual deployment of wireless nodes. On the other hand, the existing wireless planning tools only focus on office environments such that it is hard to plan IWLANs due to the larger problem size and the deployed IWLANs are vulnerable to prevalent shadowing effects in harsh industrial indoor environments. To fill this gap, this paper proposes an over-dimensioning model and a genetic algorithm based over-dimensioning (GAOD) algorithm for deploying large-scale robust IWLANs. As a progress beyond the state-of-the-art wireless planning, two full coverage layers are created. The second coverage layer serves as redundancy in case of shadowing. Meanwhile, the deployment cost is reduced by minimizing the number of access points (APs); the hard constraint of minimal inter-AP spatial separation avoids multiple APs covering the same area to be simultaneously shadowed by the same obstacle. The computation time and occupied memory are dedicatedly considered in the design of GAOD for large-scale optimization. A greedy heuristic based over-dimensioning (GHOD) algorithm and a random OD algorithm are taken as benchmarks. In two vehicle manufacturers with a small and large indoor environment, GAOD outperformed GHOD with up to 20% less APs, while GHOD outputted up to 25% less APs than a random OD algorithm. Furthermore, the effectiveness of this model and GAOD was experimentally validated with a real deployment system.

Keywords: Genetic algorithms, high performance computing, scalability, wireless networks, network planning, harsh industrial environment

## 1. Introduction

Wireless technologies are penetrating to factories, due to the advantages that they possess over cabled technologies, such as mobility, flexibility, coverage over hard-to-reach locations, as well as lower installation and maintenance cost. For instance, in some factories, WiFi is connected to the network on the shop floor (especially industrial Ethernet). This enables management devices, such as laptops and handhelds, to be temporarily connected to the industrial systems, and further facilitates onsite operators and factory managers to reconfigure the control software and change the operating parameters (Cena, Seno, Valenzano, & Zunino, 2010). Furthermore, wireless technologies play as an indispensable role (X. Gong, et al., 2016) in the concept of factories of the future (FoF), such as Industry 4.0 and Industrial Internet.

Nevertheless, an industrial indoor environment is harsh for radio propagation (X. Gong, et al., 2016; Tanghe, Gaillot, Liénard, Martens, & Joseph, 2014), compared to office environments which most wireless planning research focuses on. It is dominated by various metal or steel objects, such as production machines, storage racks, materials (steel bars, metal plates, etc.), and vehicles (automated guided vehicles or AGVs, cranes and forklifts). These obstacles shadow the radio propagation and cause coverage holes on desired areas. According to (Johan, Mikael, Tomas, Krister, & Mats, 2013), the steel, metal, and rotating machinery often cause a path loss (PL) as high as 30 - 40 dB. This jeopardizes stable wireless connection of personnel and machines on the shop floor or in the warehouse. Consequently, only one coverage layer provided by the existing wireless planning approaches (Liao, Kao, & Li, 2011; Liao, Kao, & Wu, 2011) is vulnerable to these shadowing effects.

Next to this, the PL in industrial indoor environments can be described by a one-slope PL model at 900, 2400 and 5200 MHz (Tanghe, et al., 2008). In (Plets, Joseph, Vanhecke, Tanghe, & Martens, 2011, 2012), this model further consists of a distance loss, a cumulated wall loss, and an interaction loss. The excellent correspondence between predictions and empirical measurements on the deployed network demonstrates the general applicability of a one-slope model for precise yet simple coverage prediction. This model is contrast to the Monte Carlo method (Yoon & Kim, 2013) which requires extensive computation resource and specific speedup measures. It can then be used to calculate coverage for wireless local area network (WLAN) planning (Goudos, Plets, Liu, Martens, & Joseph, 2015; N. Liu, Plets, Goudos, Martens, & Joseph, 2014; Plets, et al., 2011, 2012; Wölfle, Wahl, Wertz, Wildbolz, & Landstorfer, 2005).

Despite these achievements in WLAN planning, large-scale industrial WLAN (IWLAN) deployment has rarely been investigated in literature. For instance, the warehouse of a typical car manufacturer in Belgium measures 83,000 m$^2$. If the grid size is one meter, there are then 83,000 candidate locations for placing an AP. Most of the wireless network planning research is only limited in a small or medium scale environment, varying from several hundreds of square meters to several thousands of square meters (N. Liu, Plets, Vanhecke, Martens, & Joseph, 2015). A large building floor of 12,600 m$^2$ was considered in (Jaffres-Runser, Gorce, & Ubeda, 2006) for WLAN planning. But only 258 candidate AP locations and a dozen of APs were involved, which significantly reduces the actual complexity of the problem. A similar simplification can be observed in (Abdelkhalek, Krichen, & Guitouni, 2015; X. Liu, 2015; Mukherjee, Gupta, Ray, & Wettergren, 2011), which only enables optimization at a small or medium scale. It is challenging to perform optimization at a large scale because of the significantly increased computation resource and the stricter requirement for efficient algorithm design.

Additionally, recent wireless planning research focuses on WSN planning rather than on WLAN planning, regarding the deployment cost. A WSN often contains 10 - 1000 cheap sensor nodes (Mukherjee, et al., 2011). If one sensor costs 10 €, the deployment cost can surpass 10,000 €. As a result, the large-scale property of a WSN still makes it economically meaningful to perform WSN planning (Gupta, et al., 2015; X. Liu, 2015; Rebai, et al., 2015). Nonetheless, for a dense IWLAN, the total deployment cost is far more than ignorable due to both the much higher price of an industrial AP and the large scale. For instance, the total cost of one Siemens® Scalance W788-2 M12 AP is more than 1550 €, including the necessary accessories such as six antennas, one power cable, one power supply box, one connector, etc. Then 100 APs of the same type will cost more than 155,000 € in dense deployment, without even considering the labor cost and other engineering costs. Therefore, it is also of economic significance to minimize the IWLAN deployment cost. This significance is even enhanced when redundant APs are deployed for improving robustness, which is a prevalent WSN deployment strategy (Chen, et al., 2015).

To fill these gaps, this paper makes fourfold contributions. (1) Compared to only one coverage layer in a small office environment in literature, this papers investigates an over-dimensioning (OD) problem where two full coverage layers can be planned in a large harsh industrial indoor environment. An empirical one-slope PL model is utilized for precise yet simple coverage calculation and additionally considers the shadowing effects of three-dimensional (3D) obstacles. (2) An efficient genetic algorithm based OD (GAOD) algorithm is proposed for solving this OD problem by minimizing the number of APs that are deployed. The GA in GAOD is tailored such that the GA search can still be effective and efficient with a large number of APs and candidate locations. (3) A greedy heuristic based OD (GAOD) is further introduced, which serves as a benchmark algorithm for the GAOD. (4) This GAOD is both experimentally validated and numerically demonstrated, in comparison to most wireless planning literature that only has numerical experiments without any real deployment.

The rest of this paper is organized as follows. Sect. 2 mathematically formulates the OD problem. Sect. 3 and Sect. 4 presents GHOD and GAOD, respectively. Sect. 5 experimentally validates this model and GAOD. Sect. 6 numerically evaluates the performance of GAOD in two vehicle manufacturers' indoor environments, standing for a small and large industrial indoor environment, respectively. Sect. 7 draws conclusions.

## 2. Problem formulation

The OD problem is to minimize the number of deployed industrial APs, under the constraints of two full coverage layers in a target industrial indoor environment and an inter-AP separation longer than a limit distance. Signal blockages caused by dominant 3D metal are considered in the PL calculation. APs are assumed to be of the same type for heterogeneous planning. A solution to the OD problem is denoted by $\vec{l}$, which is a vector of AP 2D locations.

The second coverage layer serves as redundancy against coverage holes on the first layer that are potentially caused by dominant metal. If two APs are placed quite next to each other, they are very likely to be simultaneously shadowed by the same metal. To make the OD solution $\vec{l}$ robust against shadowing effects of the metal, a minimal inter-AP separation ($d_{AP\min}$), i.e., the minimal distance between any two APs, is thus a necessary constraint for this problem.

A target rectangular environment is 2D, i.e., horizontal and vertical. It is represented by two extreme 2D points, i.e., the upper left point (*xMin*, *yMin*) and the bottom right (*xMax*, *yMax*). It is discretized into $gs \times gs$ small grids, where *gs* is the grid size that is preset as an input of the model. Each grid point (GP) is represented by the upper-left point, and denoted as $gp_g$, where *g* is a unique index for each GP. A lexicographical order is applied to all the GPs, i.e.,

$$(x0, y0) < (x1, y1) \Leftrightarrow x0 < x1 \vee (x0 = x1 \wedge y0 < y1) \tag{1}$$

A target environment is thus described by a set of ordered GPs, which is denoted as $\Omega$. The GP index *i* within $\Omega$ starts from one, corresponding to the extreme point (*xMin*, *yMin*) of the rectangular environment. It increases one by one following the lexicographical order, until reaching $|\Omega|$, the total number of GPs. Then the set of GPs is denoted by $I = \{1, 2, ..., |\Omega|\}$. The following formula is used to determine the size of $\Omega$:

$$|\Omega| = ceil[(xMax - xMin)/gs] \times ceil[(yMax - yMin)/gs] \tag{2}$$

A receiver (Rx) is placed on each GP except the GPs where APs are placed. The received power on the downlink is considered in coverage calculation. Different physical bitrate requirements of an Rx have different requirements on the lowest received power, named threshold (*THLD*) hereafter. The quantified relation can be found in (X. Gong, et al., 2016). A GP is covered by an AP if the received power of the Rx on that GP is higher than or equal to the threshold. The coverage of an AP is hence represented as the GPs that are covered by this AP.

The maximal transmit power $TP_{\max}$ of an AP is considered, which is a natural way to help minimizing the number of over-dimensioned APs. All APs are of the same type. They can be placed anywhere within the target environment. Three assumptions are further made on the environment. (1) It has no previously installed APs. (2) It is so large that multiple APs are needed for even one complete layer of coverage. (3) It is empty and the shape is convex such that there is always the line-of-sight (LoS) propagation between two locations. These assumptions are reasonable, since a shop floor and a warehouse are often a large rectangular hall. The shadowing effects of dominant obstacles are considered in the PL calculation. There are $|\vec{l}|$ APs in an OD solution $\vec{l}$. The AP set is denoted as $J = \{1, 2, ..., |\vec{l}|\}$, also following a lexicographical order.

Without loss of generality, a one-slope PL model (Tanghe, et al., 2008) is used to calculate power loss between the AP Tx power and the received power of an Rx, additionally considering the signal blockages caused by dominant metal:

$$PL(d_{ij}) = PL0 + 10n\log_{10}(d_{ij}) + OL_{ij} + \xi \tag{3}$$

where *PL*0 (in dB) is the PL at the distance of one meter, *n* is the PL exponent which is a dimensionless parameter indicating the PL increase with the distance, $d_{ij}$ is the distance (in m) between the Rx placed on the *i*-th GP and the *j*-th AP, $OL_{ij}$ is the total obstacle loss (in dB) caused by the metal obstacles that block the line between the Rx placed on the *i*-th GP and the *j*-th AP, and $\xi$ (in dB) is the deviation between the measurement and model, which is attributable to shadowing.

Obstacle locations are assumed fixed in an environment. The deviation $\xi$ in Eq. (3) follows a Gaussian distribution, with a mean of zero and a standard deviation $\sigma$. The gain and margin are considered in the link budget calculation to be more realistic, which is not taken into account in (Siomina, Värbrand, & Yuan, 2007). The total gain *G* (in dB) is the sum of the AP transmitter's gain and the Rx's gain. The margin *M* (in dB) is the sum of shadowing margin, fading margin and interference margin.

The OD model is formulated in Eqs. (4-12). It is considered as large scale if the target industrial indoor environment has a large size (> 10000 m$^2$) and *gs* is small (within several meters). Otherwise, it is considered as small scale.

$$objective: \min_{(x_j, y_j) \in \Omega,\ j \in J} \left( \left| \vec{l} \right| \right) \tag{4}$$

s. t.:

$$\sum_{j=1}^{N_{AP}} \alpha_{ij} \geq 2, \forall i \in I \tag{5}$$

$$d_{\max} = 10^{\left( \frac{TP_{\max} + G - M - THD - PL0}{10n} \right)} \tag{6}$$

$$OL_{ij} = \sum_{k=1}^{N_O} \beta_{ij}^k \cdot OL_k, \forall i \in I, \forall j \in J \tag{7}$$

$$\alpha_{ij} = \begin{cases} 1, \text{if } P_{\max} + G - M - PL(d_{ij}) \geq THLD \\ 0, \text{otherwise} \end{cases}, \forall i \in I, \forall j \in J \tag{8}$$

$$\beta_{ij}^k = \begin{cases} 1, \text{if the } k\text{-th metal blocks the line between the } i\text{-th GP and the } j\text{-th AP} \\ 0, \text{otherwise} \end{cases}, \forall i \in I, \forall j \in J, \forall k \in \{0, 1, ..., N_O\} \tag{9}$$

$$d_{jj'} \geq d_{AP\min}, \forall j, j' \in J, j \neq j' \tag{10}$$

$$0 < d_{AP\min} < d_{\max}/2 \tag{11}$$

$$xMin \leq x_j \leq xMax, yMin \leq y_j \leq yMax, \forall j \in J \tag{12}$$

Eq. (4) is the objective function for OD. The objective is to minimize the number of APs ($|J|$) that are over-dimensioned. The variable is the 2D location of each AP. The output of this objective function is a vector of APs that are over-dimensioned in a target industrial indoor environment. Implicitly, the AP number is unknown and can vary for a fixed OD problem. The rest is the constraints of this OD problem.

Eq. (5) requires that each GP should be covered by at least two APs. Eq. (6) calculates the maximal distance $d_{\max}$ (in m) that an AP can potentially cover, by having the maximal transmit power on the AP side and the threshold on the Rx side, and without any obstacle blocking the line of this Tx-Rx pair. Eq. (7) calculates the total metal obstacle loss (in dB) for the pair of *i*-th GP and *j*-th AP.

Eq. (8) defines the logical blockage variable $\beta_{ij}^k$ for the *i*-th GP, *j*-th AP, and *k*-th metal obstacle. If the *k*-th metal obstacle blocks the LoS propagation between the *j*-th AP and *i*-th GP, it equals one. Otherwise, it equals zero. Eq. (9) defines the logical coverage variable $\alpha_{ij}$ for all pairs of GP-AP. If the *i*-th GP is covered by the *j*-th AP, it is one. Otherwise, it is zero.

Eq. (10) forces that any intra-AP separation should not be shorter than the preset limit distance $d_{AP\min}$. Eq. (11) sets the lower and upper bounds of $d_{AP\min}$. Eq. (12) indicates where APs can be placed: inside the rectangle target environment or just on the boundaries (i.e., side walls).

As shown in (W.-C. Ke, Liu, & Tsai, 2011; W. C. Ke, Liu, & Tsai, 2007), it is non-deterministic polynomial complete (NP-complete) to achieve *k*-cover with minimum of nodes in grid-based networks. Complying with this condition, the above formulated problem has additional constraints of obstacle shadowing and AP separation. Therefore, this problem is NP-complete.

## 3. Greedy heuristic based over-dimensioning (GHOD)

The GHOD is inspired from the recently proposed wireless planning algorithm in (N. Liu, et al., 2015) which determines the minimal number of APs and their locations while satisfying a specified physical bitrate in an office environment. The same idea is used in GHOD, by placing APs one after another. Consequently, GHOD represents a specific heuristic for this OD problem (Sect. 2).

However, two advances have been made to adapt to this OD problem. First, it creates two coverage layers instead of one and under the additional constraint of a minimal inter-AP separation. Second, it achieves the linear-time calculation in setting up the first coverage layer, by reducing the time complexity from $O(n^3)$ to $O(n)$, where n is the size of the 2D environment.

The time complexity $O(n^3)$ in the original algorithm is introduced by the $d_{avg}$ criterion in judging the best AP location when placing each AP. This criterion calculates the average distance among all uncovered GPs. It is effective to offer a minimized number of APs for a full coverage layer.

Nevertheless, two full coverage layers should be set up in OD, which changes the context of the $d_{avg}$ criterion. Besides, $O(n^3)$ is inefficient for large-scale network planning. For instance, it took 1093 sec for setting up one full coverage layer over an environment of 200 m × 50 m at a PC with an Intel i5-3470 CPU and 8G RAM. It will then take about 6.25E5 sec (about 173.6 h) for an industrial hall of 415 m × 200 m, which is very time consuming for network planning. An actual algorithm running for the latter case was conducted. As expected, there was no result after 40 h.

To reduce the time complexity, the criterion of judging the best AP is altered in GHOD, which is introduced as follows.

**Definition 1**: *covered GPs* refer to the set of GPs that are covered by at least two APs at the maximal Tx power levels.

**Definition 2**: *once-covered GPs* refer to the set of GPs that are covered by only one AP.

**Algorithm 1** A greedy heuristic based OD (GHOD)

1. *candidateGPs* ← pick out uniformly from $\Omega$;
2. **do**
3.     *bestLocation* ← GP ∈ *candidateGPs* on which an AP has the most *new once-covered GPs*;
4.     add *bestLocation* to $\vec{l}$;
5.     remove from *candidateGPs* all GPs within $d_{AP\min}$ of *bestLocation*, including *bestLocation*;
6. **while** (not all GPs are *once-covered GPs*);
7. *candidateGPs* ← $\Omega$;
8. remove from *candidateGPs* all GPs within $d_{AP\min}$ of all APs that are placed, including GPs on which the APs are placed;
9. **do**
10.     *bestLocation* ← GP ∈ *candidateGPs* on which an AP has the most *new twice-covered GPs*;
11.     add *bestLocation* to $\vec{l}$;
12.     remove from *candidateGPs* all GPs within $d_{AP\min}$ of *bestLocation*, including *bestLocation*;
13. **while** (not all GPs are *covered-GPs*);
14. Apply the lexicographical order to $\vec{l}$; ;

**Definition 3**: *new once-covered GPs* refer to the set of GPs that are not yet covered, but can be covered by a given AP at the maximal Tx power levels.

**Definition 4**: *new twice-covered GPs* refer to the set of GPs that are covered only once, and can be covered twice by a given AP at the maximal Tx power levels.

**Definition 5**: *candidate GPs* refer to the set of GPs that are available for placing APs.

As described in Algorithm 1, GHOD establishes the two coverage layers one by one. Candidate AP locations are set up first, and then iterated for picking out the best location. A new AP is placed on the best location, and is powered on with the maximal Tx power to cover as many GPs as possible. The best location is the one on which the new AP contributes to the most new once-covered GPs. The candidate AP locations are updated and iterated again for the next

It is further assumed that one GP (x1, y1) on the edge of c0 is a *once-covered GP*, while all the other GPs within c0 are *covered GPs*. Then circle1 can be got, i.e., c1: $(x-x1)^2 + (y-y1)^2 = d_{\max}^2$. No AP is actually placed within c1 except the original AP on (x0, y0).

A critical assumption is made that all the GPs on the edge of c1 is already placed with APs without respecting the constraint of Eq. (10). By following the constraints defined by Eqs. (10, 11), circle2 is drawn around the GP (x1, y1), i.e., c2: $(x-x1)^2 + (y-y1)^2 = (d_{\max} - d_{AP\min})^2$. It is then impossible to place any other AP in the area c1-c2.

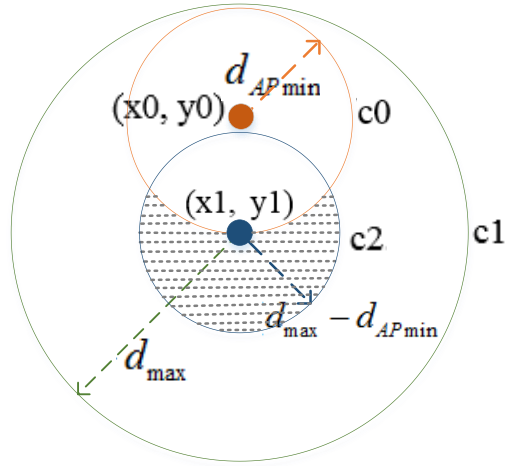

Fig. 1. Demonstration of full double coverage over the area within the minimal AP separation distance.

However, it is possible to place a new AP in the valid area c2 – (c0 ∩ c2). This valid area is also presented as the grey are in Fig. 1. It always exists due to the constraint defined by Eq. (11). As a result, the once-covered GP on (x1, y1) can always become covered GP, by placing a new AP in this valid area.

This is an extreme case. But it can still meet with property 1. Then many more loose cases exist to satisfy this property. For instance, if all the GPs on the edge of c1 are not placed with APs, the valid area to place a new AP will become larger. Furthermore, if multiple *once-covered GPs* exist regardless of the location within c0, the aforementioned process can be iterated. Therefore, property 1 is true.

Based on property 1, it can be derived that the last do-while loop (lines 9-13) in Algorithm 1 cannot be endless. Algorithm 1 is greedy, since the local best location for placing a new AP is always selected when forming up both coverage layers.

In conclusion, GHOD outputs a proper OD solution $\vec{l}$ for an OD model which is described by Eqs. (4-12).

## 4. Genetic algorithm based over-dimensioning (GAOD)

The GHOD algorithm treats an OD solution $\vec{l}$ as sequential steps, and makes the local optimal decision at each step (Sect. 3). Although it has simple time complexity of $O(n)$, it cannot guarantee a global optimal solution.

Comparatively, a genetic algorithm (GA) is one of the best-known metaheuristics in the family of evolutionary algorithms. It can give a global optimal or near-optimal solution within a reasonably short period. It has been successfully applied to solve planning problems for the industry, such as energy-cost-aware production scheduling (Gong, De Pessemier, Joseph, & Martens, 2016). GAOD is further proposed for industrial wireless planning.

*4.1 Parallel genetic algorithm*

A GA is naturally parallel and exhibits implicit parallelism, because it does not evaluate and improve a single solution but analyzes and modifies a set of solutions simultaneously (Wang, Yin, & Wang, 2009). Instead of being viewed as a mono-thread algorithm, it can be seen as a "divide and conquer" algorithm, also referring to as "map and reduce". As to the "map" phase, the data space is split into smaller and independent chunks to be processed. Once the chunks are processed, partial results are collected to form up the final result, which is the "reduce" phase.

Accordingly, parallel computing (such as multithreading) (Hwu, 2014) can be used to shorten the runtime of large-scale optimization. The substructures of a classical GA where the parallel computing can be applied include initial population generation, crossover and mutation of two individuals, and fitness calculation of a generation.

The solution encoding, initial population generation, crossover, and mutation should be defined, to link the problem-dependent information and characteristics to the general GA structure. Regarding GAOD, all the afore-mentioned operation definitions aim to minimize both memory usage and runtime for large-scale IWLAN planning.

*4.2 Solution encoding*

It requires special concern on the solution encoding for efficient optimization. Authors in (Gupta, et al., 2015) encode a wireless sensor placing solution as a vector, which contains all the candidate GPs for placing APs in a target environment. If a location is placed with a sensor, the value with the corresponding index in the vector is one. Otherwise, the value is zero. However, enormous redundancy exists in this encoding space. It is natural that the number of candidate locations is larger than or equal to the number of wireless nodes to be placed. Therefore, the candidate locations, on which no AP is finally placed, contribute nothing to the final solution. Consequently, redundancy exists in solution encoding space, which impedes the optimization efficiency. A similar concern is described in (Yoon & Kim, 2013), where the authors used a normalization method to map between genotype space and phenotype space of a GA.

To minimize memory usage and removing encoding redundancy, GAOD encodes an OD solution $\vec{l}$ as a vector that only contains the 2D locations of the over-dimensioned APs. This vector follows the lexicographical order in Eq. (1).

*4.3 Initial population generation*

The initial population generation contains *popSize* qualified individuals which are randomly generated. The purpose is to guarantee that each generated individual can satisfy all the constraints defined by the OD model, and consequently ensure the effective large-scale GA research.

Generally, it is not a prerequisite to always generate an initial individual that fully satisfy the constraints. Individuals that cannot meet with all the constraints can be later assigned with the worst fitness, and then can be eliminated throughout by elitism and roulette wheel selection (Cheng, Chuang, Liu, Wang, & Yang, 2016). Besides, these worst individuals have the opportunity to be improved through crossover and mutation.

However, an unqualified initial individual reduces the GA search efficiency, which should be considered as a crucial requirement for large-scale optimization. Otherwise, some part of the computation resource is just wasted by generating unqualified individuals as candidate solutions, and making evolution based on a mix of qualified and unqualified individuals.

An extensive generation of individuals (more than 1000 individuals) for a large-scale OD model was carried out in a computational experiment. These individuals do not necessarily satisfy the constraint of two full coverage layers that

**Algorithm 2** Initial individual generation in GAOD

1. $validGs \leftarrow \Omega$;
2. $uncoveredGPs \leftarrow \Omega$;
3. $candidateGPs \leftarrow \Omega$;
4. **do**
5.     add to $\vec{l}$ a random GP of *candidateGPs*;
6.     remove from *validGPs* all GPs within $d_{AP\min}$ of this GP;
7.     place a new AP on this GP and power it on with $TP_{\max}$;
8.     increase by one the coverage layer number of each GP that is within the coverage of this AP;
9.     remove the new *covered GPs* from *uncoveredGPs*;
10.     $candidateGPs \leftarrow validGPs \cap uncoveredGPs$;
11. **while** ($candidateGPs \neq \emptyset$);
12. **while** ($uncoveredGPs \neq \emptyset$)
13.     $centerGP \leftarrow$ the first GP in *uncoveredGPs*;
14.     $poolGPs \leftarrow$ all GPs within the $d_{AP\min} \times d_{AP\min}$ square which is centered at *centerGP*;
15.     $candidateGPs \leftarrow poolGPs \cap validGPs$;
16.     **if** ($candidateGPs == \emptyset$)
17.         $poolGPs \leftarrow$ all GPs within the $d_{AP\min} \times d_{AP\min}$ square which is centered at *centerGP*;
18.         $candidateGPs \leftarrow poolGPs \cap validGPs$;
19.     **end if**
20.     add to $\vec{l}$ a random GP of *candidateGPs*
21.     remove from *validGPs* all GPs within $d_{AP\min}$ of this GP;
22.     place a new AP on this GP and power it on with $TP_{\max}$;
23.     increase by one the coverage layer number of each GP that is within the coverage of this AP;
24.     remove the new *covered GPs* from *uncoveredGPs*;
25. **end while**
26. apply the lexicographical order to $\vec{l}$;

are defined by Eq. (5). However, this experiment could not output even one qualified initial OD solution. This in turn demonstrates the necessity of Algorithm 2.

**Definition 6**: *uncovered GPs* refer to the set of GPs that cannot be covered twice by all placed APs at the maximal Tx power level.

**Definition 7**: *valid GPs* refer to the set of GPs that are located beyond $d_{AP\min}$ of all the APs that are already placed in the environment.

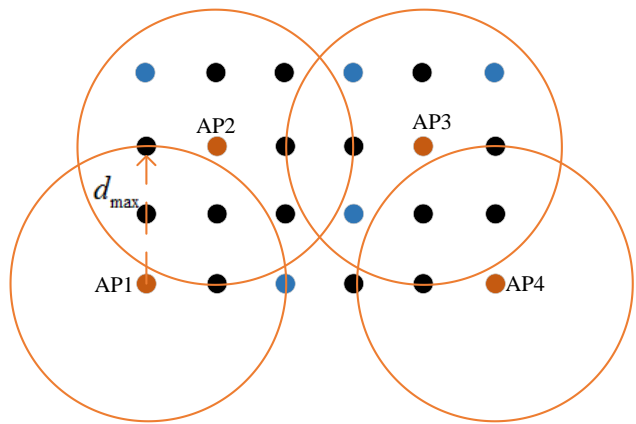

Fig. 2. Demonstration of full double coverage over the area beyond the minimal AP separation distance ($d_{max}/2$). Black dots represent the grid points on which no AP is placed. Orange dots are AP locations for the first coverage layer. Blue dots are AP locations for the second coverage layer. Large orange circle is the coverage area of an AP on the first coverage layer.

Algorithm 2 shows the method to produce a qualified initial individual. It comprises two parts after the initialization (lines 1-3): iteration 1 (lines 4-11) and iteration 2 (lines 12-25).

The initialization allows the accommodating PC to temporally allocate memory to the three variables (i.e., *validGPs*, *uncoveredGPs* and *candidateGPs*), instead of storing them locally at each individual. This is because the number of GPs in an industrial indoor environment can be huge. It may take up a significant amount of memory to represent all the GPs. Consequently, the GA search may be threatened by a lack of memory.

Iteration 1 places APs one by one on GPs that are covered less than twice and beyond $d_{APmin}$ of all APs that are already placed. It is greedy in the sense that the number of *uncovered GPs* rapidly decreases one loop after another.

But it does not have to be strictly greedy like GHOD, since this will not guarantee the global optimum while rising the computation burden. Furthermore, it cannot be endless. This is demonstrated by property 2.

**Property 2**: by placing APs of the same type one by one, all the *valid GPs* can be covered at least twice, i.e., iteration 1 (lines 4-11) of Algorithm 2 cannot be endless.

**Proof**: given the constraint in Eq. (11), this property can be bounded by two extreme cases, i.e., $d_{APmin}=0$ and $d_{APmin}=d_{max}/2$. If $d_{APmin}=0$, then two APs can be placed on the same GP for double full coverage of the same area.

If $d_{APmin}=d_{max}/2$, the environment is sure to have the first coverage layer. One specific case is demonstrated in Fig. 2, where $d_{max}$ is 2*gs*. The first full coverage layer can be formed up by AP1, AP2, AP3 and AP4. Then, the environment can be fully covered for the second time, by five additional APs, represented as blue dots in Fig. 2. Obviously, all the placed APs are beyond $d_{APmin}$ (*gs*). This process can be iterated until reaching the second full coverage layer. Although the two extreme cases are tight, property 2 is true. There are many more cases between the two bounds that can meet with property 2. Therefore, property 2 is true.

Iteration 2 intends to cover all the *uncovered GPs* are located within $d_{APmin}$ of the APs that are already placed. The candidate GPs are picked out from a defined rectangular area, which is centered at the first element of the set of

**Algorithm 3** Crossover of GAOD

1. *xMin* ← max (minimal horizontal coordinates of all APs in indiv1 and indiv2);
2. *xMax* ← min (maximal horizontal coordinates of all APs indiv1 and indiv2);
3. *xCrossover* ← a random coordinate $\in$ [*xMin*, *xMax*);
4. chop graphically *indiv1* and *indiv2* into two parts along the same vertical line *xCrossover*, respectively
5. *newIndiv1* ← 1$^{st}$ part of *indiv1* + 2$^{nd}$ part of *indiv2*;
6. *newIndiv2* ← 1$^{st}$ part of *indiv2* + 2$^{nd}$ part of *indiv1*;
7. **for** *indiv* $\in$ {*newIndiv1*, *newIndiv2*}
8.    remove APs that are within $d_{AP\min}$ of any APs in the area $xCrossover - d_{AP\min} \leq x \leq xCrossover + d_{AP\min}$;
9.    *uncoveredGPs* ← $\Omega$;
10.    **for** AP $\in$ {APs that remain in *indiv*}
11.       increase by one the coverage layer number of each GP within the coverage of the new AP;
12.       remove all new *covered GPs* from *uncoveredGPs*;
13.    **end for**
14.    **if** (*uncoveredGPs* $\neq \emptyset$)
15.       *validGPs* ← $\Omega$;
16.       **for** AP $\in$ {APs that remain in *indiv*}
17.         remove from *validGPs* GPs within $d_{AP\min}$ of this AP;
18.       **end for**
19.       *candidateGPs* ← *validGPs* $\cap$ *uncoveredGPs*;
20.       iteration 1 (lines 4-11) in Algorithm 2;
21.       iteration 2 (lines 12-25) in Algorithm 2;
22.    **end if**
23. **end for**
24. apply the lexicographical order to *newIndiv1* and *newIndiv2*;

*uncovered GPs*. First, it is a small $d_{AP\min} \times d_{AP\min}$ area aiming to place the new AP as close to the first *uncovered GP* as possible, while respecting the constraint defined by Eq. (10) and beyond $d_{AP\min}$ of all APs that are already placed. If this small area has no qualified GP, a large $d_{\max} \times d_{\max}$ area is then created. Given property 1, there must exist at least one GP within this large area to cover the first *uncovered GP* the second time. Consequently, iteration 2 cannot be endless, either.

Moreover, as Algorithm 2 generates a random OD solution, it can be used as a benchmark for GHOD and GAOD. An individual is $\vec{l}$. The numbers of APs in different individuals do not have to be the same, due to the randomness of

Algorithm 2. The minimization of AP number will depend on the population evolution, which is driven by crossover, mutation and elitism.

*4.4 Crossover*

The crossover operation is defined by Algorithm 3. The input is two qualified individuals (i.e., *indiv1* and *indiv2* in Algorithm 3), which are selected by the roulette wheel selection algorithm (Cheng, et al., 2016).

The crossover point is defined as a vertical line, named *xCrossover*. The horizontal coordinate of *xCrossover* is randomly selected (line 3) from the effective range calculated by lines 1-2 in Algorithm 3. This vertical line splits the rectangular environment into two rectangular subparts, i.e., the parts of which all the involved horizontal coordinates are smaller (part 1) and larger (part 2) than the randomly selected one, respectively. Then the two parts on the two individuals are swapped to get two children solutions (lines 4-6).

The minimal AP separation constraint defined by Eq. (10) may be broken after the swap. However, it is unnecessary to check over all the environment, since this can only occur within the small rectangular area around the vertical split line, i.e., $xCrossover - d_{AP\min} \leq x \leq xCrossover + d_{AP\min}$. Thereby, for speedup within this small rectangular area, if an AP is within $d_{AP\min}$ of another AP, this AP is removed from the OD solution $\vec{l}$ represented by the current child individual (line 8).

The two children solutions are then checked (lines 9-13) whether they satisfy the constraint of two full coverage layers defined by Eq. (5). If this constraint cannot be satisfied, iterations 1 and 2 in Algorithm 2 will be performed (lines 14-22, Algorithm 3). This is not costly in terms of time and memory, since after a swap, this constraint can be broken only in the small area $xCrossover - d_{AP\min} \leq x \leq xCrossover + d_{AP\min}$.

Moreover, memory-consuming variables (such as *uncovered GPs* in Algorithm 3) do not have to be locally stored in the individuals and population. All these variables are locally generated, meaning that the occupied huge memory will be immediately freed up at the end of Algorithm 3.

*4.5 Mutation*

A mutation operation produces a new qualified individual. It should be different to all the existing individuals as much as possible, because in concept mutation adds diversity to a generation and avoids a GA search to quickly converge in a single direction in the solution space.

To this end, Algorithm 4 is designed for mutation in the ODGA. It mainly consists of two steps. At step 1 (lines 1-10), additional APs of the same type are added to the target environment, while respecting the minimal AP separation constraint defined by Eq. (10). At step 2 (lines 11-18), the APs in the original OD solution are checked one after another whether an AP can be removed, while still satisfying the constraint of full double coverage defined by Eq. (5).

Similar to the former algorithms, there is no need to locally store memory-consuming variables (such as *validGPs* in Algorithm 4) in each individual. All these variables are locally generated. The huge memory taken by them will consequently be freed up at the end of Algorithm 4.

**Algorithm 4** Mutation of GAOD
1. $validGPs \leftarrow \Omega$;
2. **for** AP $\epsilon$ {original APs placed on $\vec{l}$ }
3.    remove from $validGPs$ all GPs within $d_{AP\min}$ of this AP;
4. **end**
5. $numAdditionalAPs$ = ceil($rateMutation \cdot numAllAPs \cdot 0.5$);
6. **for** $i = 1 : numAdditionalAPs$
7.    place a new AP on a random GP $\epsilon$ $validGPs$;
8.    add the new AP to the set $additionalAPs$;
9.    remove from $validGPs$ all GPs within $d_{AP\min}$ of this AP;
10. **end for**
11. **for** AP $\epsilon$ $additionalAPs$
12.    add GPs that are covered by this AP to the set $newCovGPs$;
13. **end for**
14. **for** AP $\epsilon$ {original APs placed on $\vec{l}$ }
15.    **if** (all GPs covered by this AP $\subset$ $newCovGPs$)
16.      remove the location of this AP from $\vec{l}$;
17.    **end if**
18. **end for**
19. apply the lexicographical order to the new vector $\vec{l}$;

*4.6 Additional speedup measures*

As described in the former subsections, the design of ODGA follows the idea of saving the memory and computation time as much as possible, in order to facilitate large-scale optimization. Next to this, additional specific speedup measures are taken on two types of calculations which are extensively used in ODGA.

First, $d_{\max}$ is calculated by the PL model in advance and stored as a constant, instead of repeating the same PL calculation for millions of times during the GA search. Second, Algorithms 2-4 extensively search the area that an AP can cover with $TP_{\max}$, as well as the area which is within $d_{AP\min}$ of an AP. Instead of a rude iteration of all GPs in the environment to find the qualified GPs, such a search is only restricted within the $d_{\max} \times d_{\max}$ or $d_{AP\min} \times d_{AP\min}$ rectangular area which is centered at the investigated AP.

## 5. Experimental validation

The over-dimensioning (OD) model was experimentally validated in a small open environment (nearly 10 m × 10 m) in the factory hall of an AGV manufacturer, in Flanders, Belgium.

*5.1 Configurations*

A PC accommodating all the algorithms, a measurement control system (X. Gong, et al., 2016), as well as four COTS Siemens® industrial APs (Scalance W788-2) with individual power supply were used. The WLAN coverage measurement facilities were the same as these introduced in (X. Gong, et al., 2016). They mainly include a measurement control software system, a Zotac® mini-PC as a wireless client, an AGV as the controllable mobile vehicle which carries the client on the top, and four poles with tripods to support the APs. Besides, 44-dB attenuators were applied to the four APs, to mimic a larger environment that needs four APs for double full coverage.

In total, 3745 RF power samples were collected by driving the AGV around the environment at 20 cm/s. Regression (Tanghe, et al., 2008) was applied to these data to build an empirical PL model formulated by Eq. (3). $PL_0$ is 39.87 and $n$ is 1.78. The R-squared value is 97.38%, indicating a high fitness level of the PL model, compared to the samples. The factory hall is more than 10 m high, such that it was only possible to place the four APs on the four sides of the environment with the aforementioned facilities. Besides, the AGV moves around in the environment for measurements.

**Table 1**

**Configurations of the environment and genetic algorithm (GA)**

| Wireless configurations | |
|---|---|
| Shadowing margin (95%) | 1 dB |
| Fading margin (99%) | 0 dB |
| Interference margin | 0 dB |
| AP antenna attenuation | 44 dB |
| WLAN standard | IEEE 802.11n |
| Frequency band | 2.4 GHz |
| AP height | 2 m |
| AP only on the wall? | Yes |
| Minimal AP separation ($d_{AP\min}$) | 5 m |
| Client height | 1.8 m |
| Required physical bitrate of a client | 24 Mbps |
| Required minimal sensitivity of a client | -79 dBm |
| GA configurations | |
| Population size | 100 |
| Elitism rate | 10% |
| Crossover rate | 90% |
| Mutation rate | 5% |
| Maximum iteration | 30 |

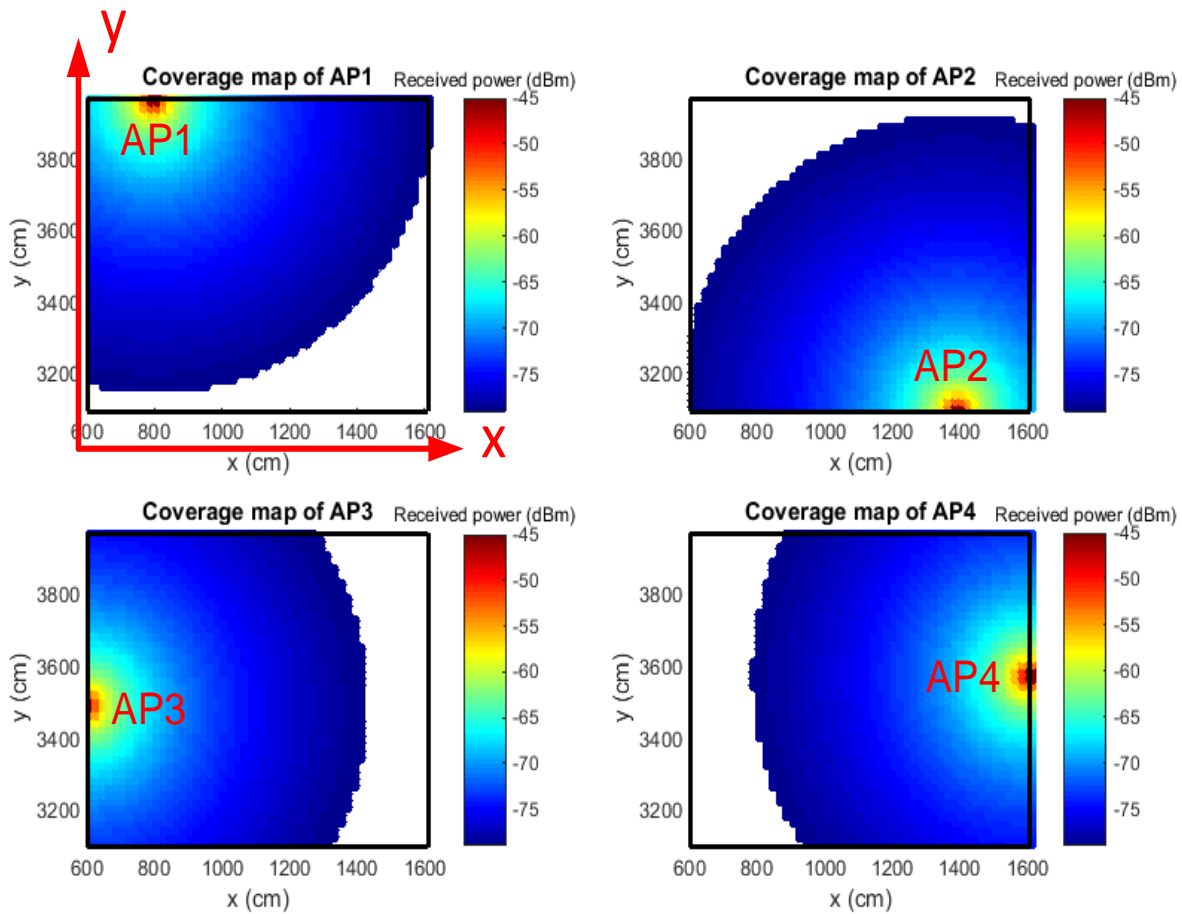

Fig. 3. Coverage of the four over-dimensioned access points (APs), which is predicted by the path loss model. The colored area is covered by an AP, while the white area is out of coverage.

Placing APs within the environment will impede the AGV's movement. Based on the constraint of AP locations defined by Eq. (12), a tighter constraint was then made in the validation: APs can be placed only on the boundary of the environment. Moreover, the minimal AP separation distance $d_{AP\min}$ was set as five meters.

Table 1 summarizes the key configurations of this validation, including theses for the environment and the genetic algorithm based over-dimensioning (GAOD).

*5.2 Results*

The obtained OD solution is illustrated in Fig. 3. The thick black lines represent the four boundary sides of the environment. Four APs are placed on the boundary of the environment, such that two full coverage layers are envisioned to be made. Each AP has the exact 2D location (in cm).

Fig. 3 also shows the coverage of each over-dimensioned AP. It consists of four subfigures. Each one illustrates the coverage of an AP. The red represents the area with high RF power, the blue represents the area with low RF power which is however not lower than the required minimal sensitivity (-79 dBm), while the while stands for the area that cannot be covered by the AP. As shown, every AP cannot fully cover the environment. However, two APs can form up a complete coverage layer by combing the respective coverage, i.e., AP1 and AP2, as well as AP3 and AP4. The minimal inter-AP distance is 5.13 m between AP1 and AP3. This is larger than the preset $d_{AP\min}$ (5 m). Therefore, this numerically demonstrates the effectiveness of the OD model and ODGA.

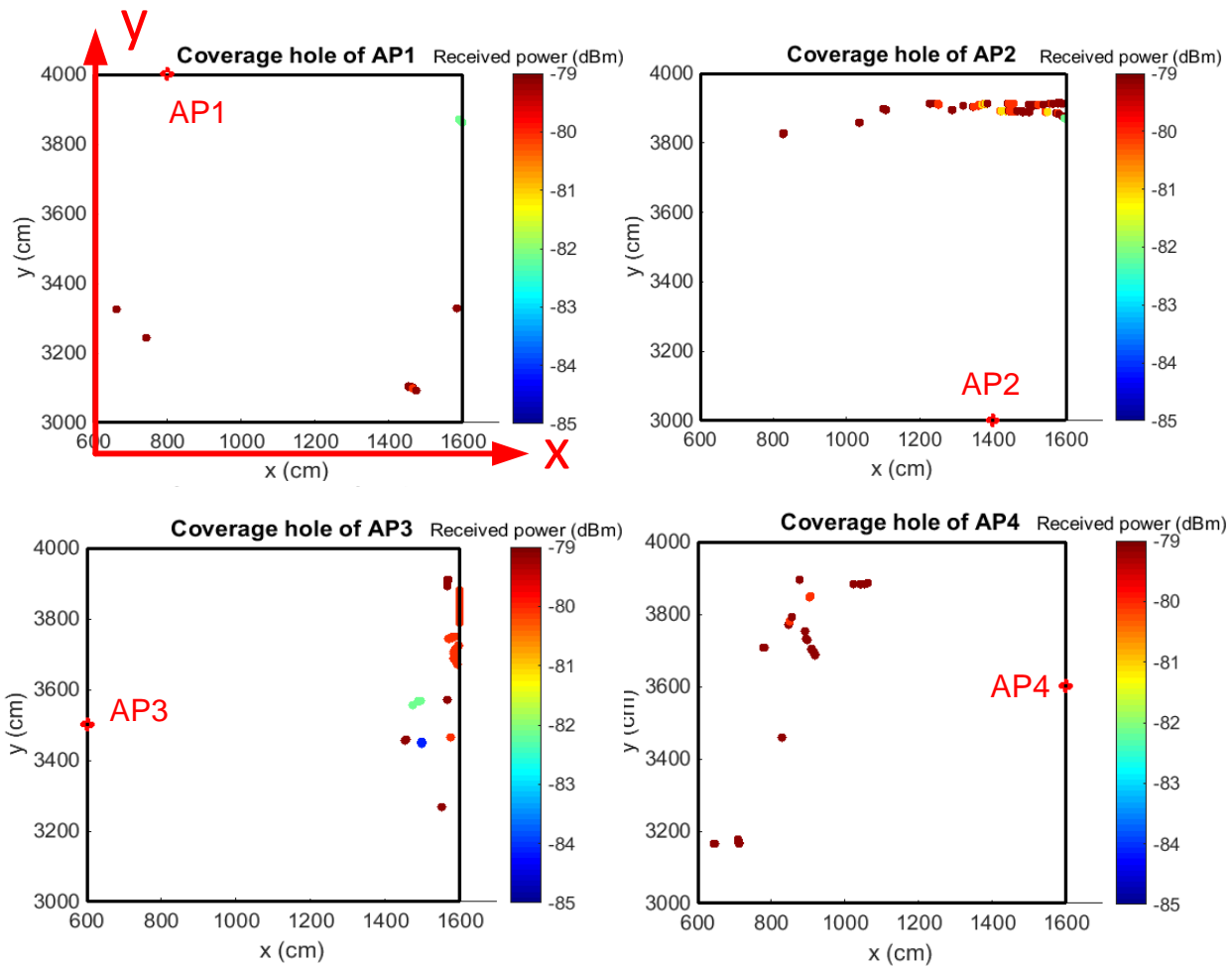

Fig. 4. Coverage hole of the four over-dimensioned access points (APs), which is measured by experiments.

The four APs were placed in the environment according to the obtained OD solution. The coverage of each AP was measured, by sweeping the entire environment with the AGV equipped with the client. Fig. 4 presents the RF power samples that are lower than the required minimal sensitivity, i.e., coverage hole of each AP. The coverage hole of an AP is always near to the boundary of the environment and on the opposite side of this AP location. The coverage power samples vary between -79 dBm and -85 dBm, which are actually below the present sensitivity (-79 dBm). Therefore, this empirically demonstrates the effectiveness of the OD model and ODGA.

## 6. Numerical experiments

The focus of numerical experiments is on the algorithmic scalability beyond the experimental validation at a small scale (Sect. 5), to adapt to the real industrial wireless deployment scale.

The models and algorithms were implemented in Java. The numerical experiments were performed on a PC running 64-bit Win7 and with an Intel i5-3470 CPU (two 3.20 GHz single-thread cores) and an 8 GB RAM.

**Table 2**

**Numerical experiment configurations**

| | |
|---|---|
| Path loss model | |
| PL0 | 39.87 dB |
| $n$ | 1.78 |
| Shadowing margin (95%) | 1 dB |
| Fading margin (99%) | 0 dB |
| Interference margin | 0 dB |
| Transmitter of an access point (AP) | |
| Height | 2 m |
| Gain | 3 dB |
| WiFi standard | IEEE 802.11n |
| Frequency band | 2.4 GHz |
| Maximal transmit power | 7 dBm |
| Only on the wall? | No |
| Receiver of a wireless client | |
| Height | 1.4 m |
| Gain | 2.15 dB |
| Required physical bitrate | 54 Mbps |
| Required minimal sensitivity | -68 dBm |
| Environment | |
| Size of the factory hall (small) | 102 × 24 m (2600 grid points) |
| Size of the warehouse (large) | 415 m × 200 m (83616 grid points) |
| Grid size ($gs$) | 1 m |
| Frequency band | 2.4 GHz |
| Antenna type | Omnidirectional |
| Minimal inter-AP separation | 5 m |
| Metal rack size | 20 m × 3 m × 9 m |
| Path loss caused by one metal rack | 7.37 dB |
| GAOD configurations | |
| Population size | 30 (small-scale environment), 100 (large-scale environment) |
| Elitism rate | 8% |
| Crossover rate | 95% |
| Mutation rate | 5% |
| Stop criterion | No improvement of the best fitness value during 10 consecutive cycles |

*6.1 Configurations*

The two industrial indoor environments under investigation are, respectively, a factory hall of an automated guided vehicle (AGV) manufacturer and a warehouse of a car manufacturer, both located in Flanders, Belgium.

The factory hall measures 102 m × 24 m. Metal racks are placed inside for component storage. AGVs of varying sizes are placed usually without moving and waiting for integration, maintenance, or shipment. Wide WiFi coverage is needed for AGV communication and Internet access of onsite laptops.

The warehouse measures 415 m × 200 m. Metal racks are placed inside, at a height of nine meters. Wooden boxes that contains metal components are placed on the racks. Wide WiFi coverage is required to support voice picking. The pickers are equipped with microphones and earphones. They communicate with the control center via WLANs to pick up and place a stuff at a specific location.

Mapping to the OD model, a metal rack in both cases is an obstacle that potentially causes evident shadowing effects to radio propagation. In the following experiments, an obstacle measures 20 m × 3 m × 9 m. It can be placed either horizontally (the length side is parallel to the length side of the environment) or vertically (the length side is parallel to the width side of the environment). The direction and location of an obstacle are randomly and uniformly generated in the environment. The number of racks is an input of the OD model. The GPs that are taken up by obstacles are not considered for the PL calculation.

The experiment-related parameters are shown in Table 2, including the PL model, the AP transmitter, the receiver, the environment and the GAOD. All APs are powered on with maximal transmit power ($TP_{max}$). The grid size (*gs*) is set as one meter. It is within the distance of 10 wave length (≈ 1.2 m) at 2.4 GHz radio frequency band, meaning that the PL within this distance can be considered constant without sacrificing the precision of PL calculation. The two parameters *PL*0 and *n* for the one-slope PL model were same as these in Sect. 5. The PL caused by a metal rack (7.37 dB) was the mean value of the measured PL data.

Furthermore, two benchmark algorithms are used to evaluate the performance of GAOD: GHOD and random placement (Algorithm 2). The latter is a common method to deploy wireless sensors (Jain & Ramana Reddy, 2015).

*6.2 Results in a small-scale empty environment*

GHOD and GAOD were first performed in the small-scale empty environment (i.e., factory hall of the AGV manufacturer), by making loose the constraint in Eq. (7) such that no metal obstacles exist (i.e., $N_O = 0$). The performance metrics of both algorithms are shown in Table 3. Both algorithms satisfy the constraints of two full coverage layers and $d_{AP\min}$ in the target factory hall, i.e., Eqs. (5, 10). However, GAOD outputs one AP less to solve the same OD model, and is 2.7 times faster as GHOD.

In the solution output by GHOD, the number of GPs that are covered by at least three APs is 3.3 times as the same type of number in GAOD. This unveils an intrinsic characteristic of GAOD: it essentially minimizes the number of GPs that are covered by more than two APs, while ensuring each GP is covered by at least two APs.

Moreover, 230 random OD solutions are generated by using Algorithm 2. As indicated in Table 3, on average five APs are needed with a standard deviation of one AP. This means that the median case corresponds to GHOD, and the

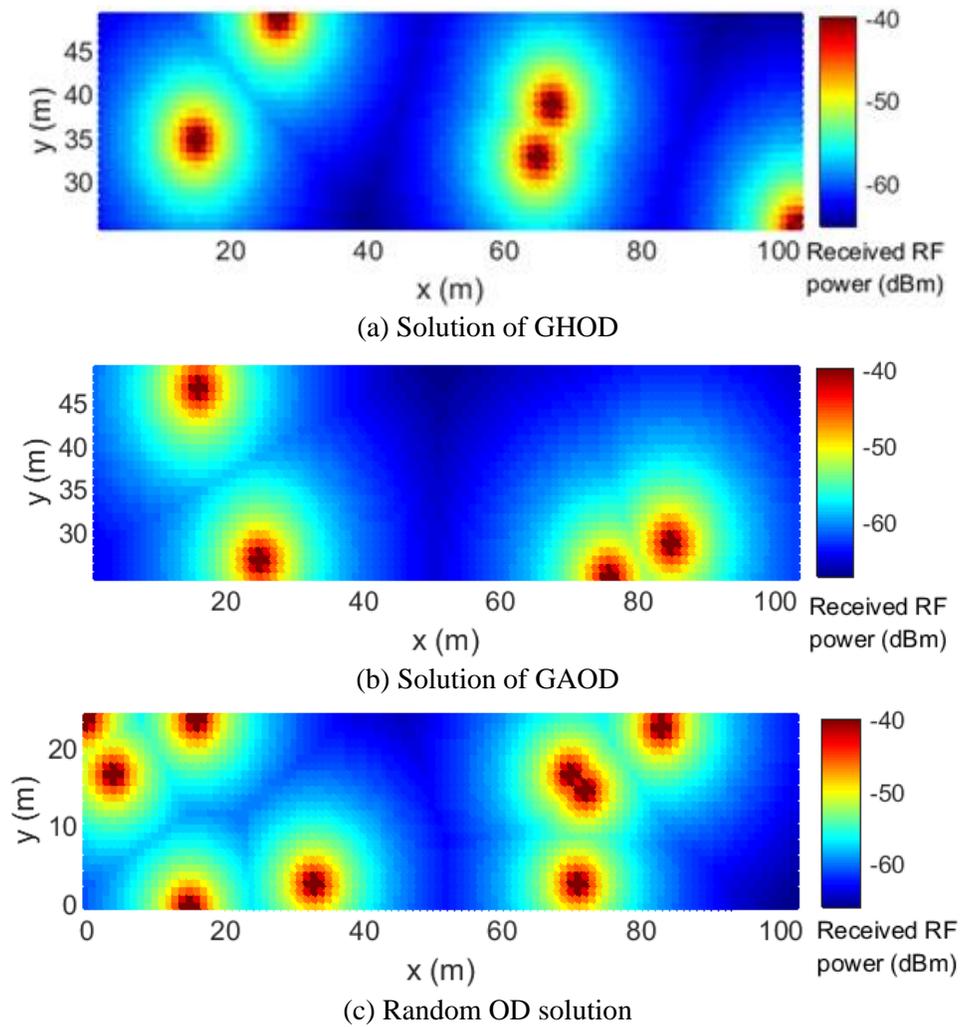

Fig. 5. Over-dimensioning solution comparison in a small-scale empty environment.

worst case outputs 50% more APs than GAOD, which accordingly leads to about 50% more AP deployment cost than GAOD. The time to generate a random solution is negligible. This is normal since a random instance only needs to meet with the two fundamental constraints defined by Eqs. (5, 10), without any optimization effort. Overall, both

**Table** 3

**Algorithm performance in empty environments**

| Performance metric | GHOD | | GAOD | | Random OD (mean/deviation) | |
|---|---|---|---|---|---|---|
| | S[a] | L[b] | S | L | S | L |
| Number of all APs | 5 | 81 | **4** | **75** | 5/1 | 85/3 |
| Runtime (sec) | 8 | 2633 | 3 | 19789 | 0/0 | 522/142 |
| Percentage of GPs covered at least twice | 100 | 100 | 100 | 100 | 100/0 | 100/0 |
| Any AP separation within $d_{AP\min}$? | No | No | No | No | No | No |
| Percentage of GPs covered more than twice | 70 | 93 | **21** | **84** | 65/18 | 91/2 |

[a]Small-scale environment. [b]Large-scale environment.

proposed algorithms can give effective OD solutions, while GAOD is superior to GHOD on the small scale in terms of AP number that is output and computation time.

As a comparison, the OD solutions output by GHOD, GAOD and random generation (Algorithm 2) are shown in Fig. 5. The x and y axes are the horizontal and vertical sides of the factory hall under investigation, respectively. The highest received power of each GP is visualized. High power is represented by red, while low power is indicated by blue. As a result, the over-dimensioned APs are represented by red dots. Fig. 5 evidently shows that GAOD outputs the least APs, while the random generation outputs the most APs within the same environment. In the solution of GAOD, APs tend to be evenly distributed over the environment. In the solution of random generation, APs however tend to be clustered, which also reveals why more APs are needed for satisfying the same constraints of the same OD model.

Fig. 5 also serves as a heat map for network managers and plan managers. It vividly reveals the coverage of the whole industrial indoor environment. The minimal received power on this map is -67 dBm, which is higher than the threshold -68 dBm (Table 2). The minimal inter-AP separation is 9.8 m, which is larger than $d_{AP\min}$ (5 m). This proves that the output OD solution satisfies the essential constraint in Eq. (10).

*6.3 Results in a large-scale empty environment*

GHOD and GAOD were then carried out in the large-scale empty environment (i.e., warehouse of the car manufacturer), by making loose the constraint in Eq. (7) such that no metal obstacles exist (i.e., $N_O = 0$). The performance metrics of are presented in Table 3. Both algorithms meet with the two essential constraints, i.e., double full coverage, and any AP beyond $d_{AP\min}$ of all the other APs. Most importantly, in terms of the optimization objective, GAOD outputs six APs less. This roughly corresponds to 7.4% reduction of the network deployment cost.

GHOD is 7.5 times faster as GAOD. This is inverse to the phenomenon revealed in the former case. It is explained by the $O(n)$ time complexity of GHOD, of which the fast performance shows up when the problem size grows rapidly. Nevertheless, the time taken is not a crucial factor for wireless planning, since the planning is performed only once or at a very low frequency. Besides, the time (5.5 h) taken by GAOD is considered acceptable, and is significantly improved compared to 173.6 h in the former experiment of running the $d_{avg}$ criterion (Sect. 3).

In the OD solution given by GHOD, 9% more GPs are covered by at least three APs. This again demonstrates the aforementioned GAOD's intrinsic characteristic of global optimization.

Furthermore, 380 random solutions are generated. As shown in Table 3, 85 APs in average are needed with a standard deviation of three APs. This mean AP number is 4.9% and 13.3% larger than the number of APs output by GHOD and GAOD, respectively. The best case in the random OD solution has 82 APs, which is still worse than the performance of GHOD and GAOD, due to the increasing AP deployment cost. The time for establishing an OD solution is much shorter than GHOD and GAOD, which is the same phenomenon as in the small scale due to the same reason. The percentage of GPs that are covered for more than twice is higher than GAOD, and is at the similar level of GHOD.

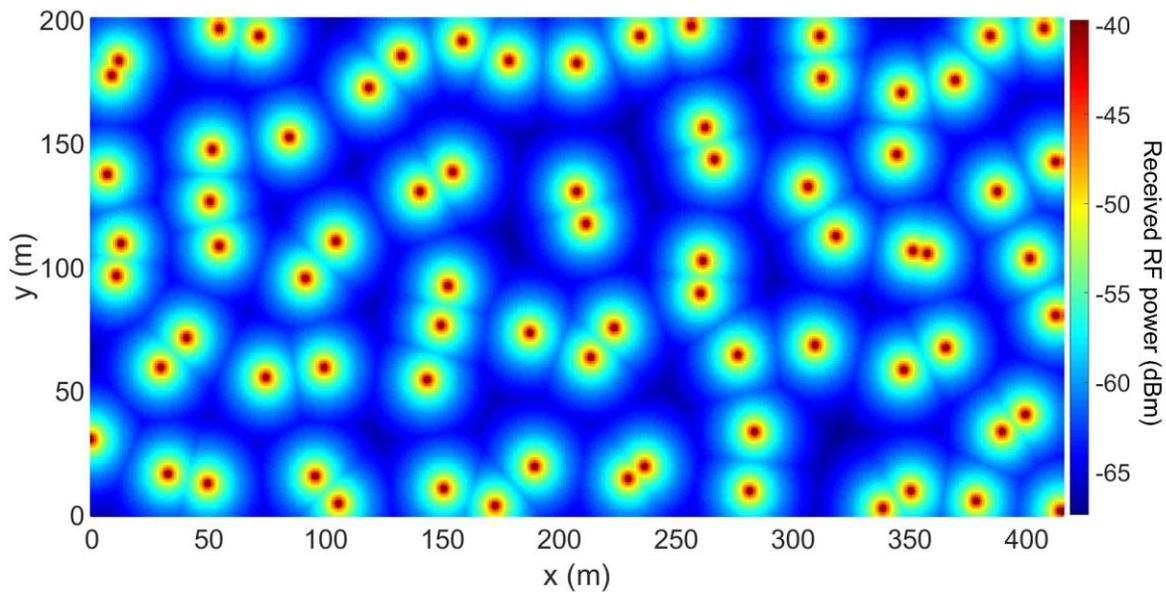

Fig. 6. Over-dimensioning solution given by GAOD in a large-scale empty environment.

The OD solution given by GAOD is further visualized in Fig. 6. The minimal received power on this heat map is -67.42 dBm, which is higher than the threshold. The minimal inter-AP separation among all the over-dimensioned APs is 6 m, which is higher than $d_{AP\min}$ (5 m).

*6.4 Results in obstructed environments*

The GHOD and GAOD algorithms were further performed in the small-scale and large-scale obstructed environments, respectively. One and ten metal racks (Table 2) were randomly placed in the small-scale and large-scale environments, respectively. In total, 330 and 200 random OD solutions were generated in the small-scale and large-scale environment, respectively.

All the obtained OD solutions meet with all the constraints of the model. Table 4 lists the other key performance metrics for comparison. All the solutions are obtained within a reasonable time span, regarding the problem size and context of wireless planning. However, GAOD can output the least APs in both small and large environments that have metal racks. A random solution outputs 40% more APs in average and 60% more APs in the worst case, compared with GAOD. The number of APs that are over-dimensioned by GHOD is median, compared with the other two algorithms.

Moreover, the percentage of GPs that are covered more than twice exhibits a similar performance trend: GAOD achieves the lowest percentage, while a random solution and GAOD have an evidently higher percentage. In a small-scale obstructed environment, the percentage of GAOD is 22% less than that of GHOD, and is in average 9% less than that of a random solution. In a large-scale obstructed environment, the percentage of GAOD is 6% less than that of GHOD, and is in average 5% less than that of a random solution. The gap between GAOD and the other two algorithms reduces in a large-scale environment.

**Table 4**

**Algorithm performance in obstructed environments**

| Algorithm | | Number of APs | Runtime (sec) | % of GPs covered at least twice |
|---|---|---|---|---|
| GHOD | S[a] | 6 | 2 | 88 |
| | L[b] | 91 | 23956 | 91 |
| GAOD | S | **5** | 7 | **66** |
| | L | **83** | 34028 | **85** |
| Random OD (mean/deviation) | S | 7/1 | 0/0 | 75/11 |
| | L | 92/2 | 543/54 | 90/1 |

[a]Small-scale environment. [b]Large-scale environment.

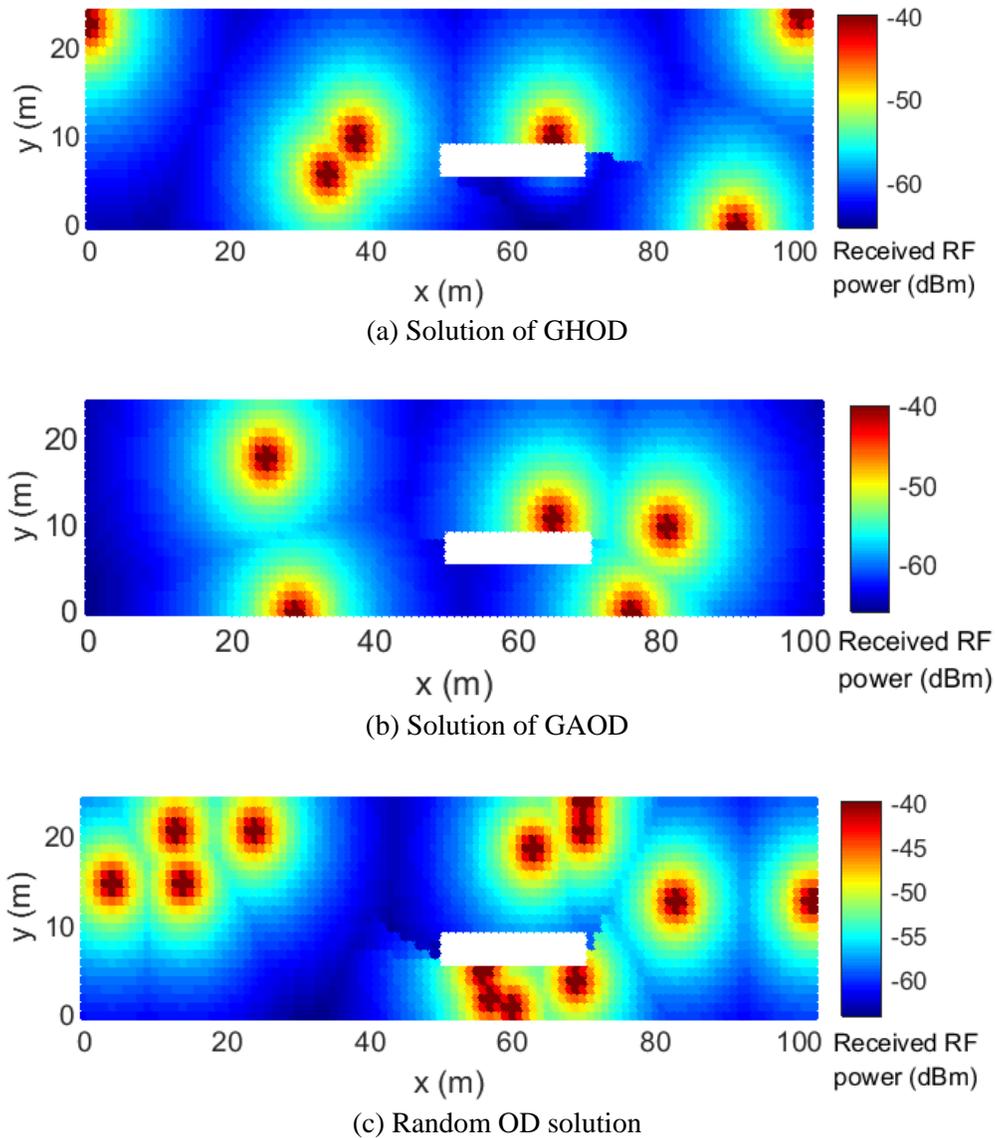

(a) Solution of GHOD

(b) Solution of GAOD

(c) Random OD solution

Fig. 7. Over-dimensioning solution comparison in a small-scale obstructed environment (the white rectangle represents a metal rack).

One important reason for this phenomenon is that more metal racks are placed in the large-scale environment, which causes more shadowing effects. Consequently, more additional APs are needed to specifically tackle these shadowing effects, while increasing the coverage layer number of the other GPs that are already covered twice.

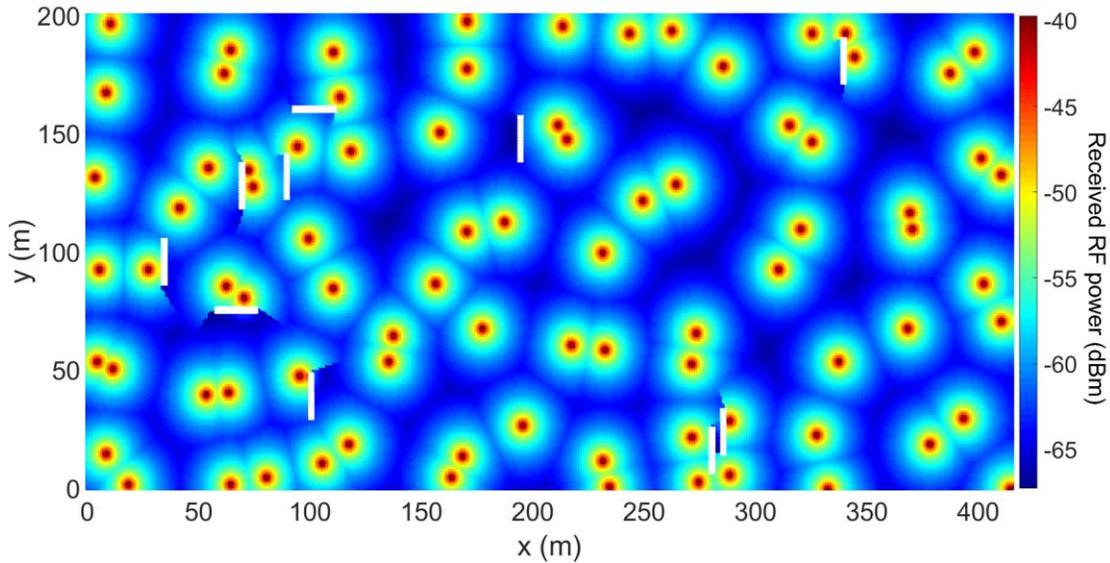

Fig. 8. Over-dimensioning solution given by GAOD in a large-scale obstructed environment (the 10 white rectangles represent 10 metal racks).

A performance comparison is made between the OD solutions which are output by the same algorithm in the same environment without and with metal, respectively. In the small-scale environment, more APs are output by the three algorithms if dominant metal is present. The AP number increasing rate of a random solution is the highest, at 40% in average. This is explained by a lack of effective optimization measures in random solution generation. In reverse, this demonstrates the effectiveness of GHOD and GAOD, in terms of minimizing the AP number. In the large-scale environment, similarly, more APs are needed under the presence of dominant metal. The AP number increasing rate of the three algorithms is on the same level around 10%, while the rate of GHOD is the highest, at 12%. This shows that the optimization performance of GHOD is worse than GAOD.

The coverage map of the OD solution in the small-scale obstructed environment output by GHOD, GAOD, and random generation, is further visualized in Fig. 7. The superiority of GAOD is clearly demonstrated, in terms of minimizing the AP number while meeting with all the constraints of the OD model.

The coverage map of the OD solution in the large-scale obstructed environment output by GAOD is further presented in Fig. 8. The 10 metals racks that are randomly generated are represented by 10 white rectangles. It is observed that APs tend to cluster around metal racks. This reveals the intrinsic property of GAOD when dealing with shadowing effects of dominant metal: additional APs are actually placed to tackle the additional PL that is caused by these shadowing effects.

## 7. Conclusion

Although wireless technologies are penetrating to the manufacturing industry, the existing research on wireless local area network (WLAN) planning is still limited in small office environments. Consequently, the one coverage layer provided by these WLAN planning approaches is vulnerable to the shadowing effects of prevalent metal obstacles in harsh industrial indoor environments. To fill this gap, this paper investigates an over-dimensioning (OD) problem where two full coverage layers can be created at a large industrial scale for robust industrial wireless coverage. Although the second coverage layer serves as redundancy against shadowing, the deployment cost can be reduced by minimizing the number of access points (APs), while respecting the practical constraint of a minimal inter-AP spatial separation.

A genetic algorithm based OD (GAOD) algorithm is proposed to solve this problem. To enable large-scale industrial WLAN (IWLAN) planning, solution encoding, initial population generation, crossover and mutation are designed, such that the required computation time and memory are minimized. A greedy heuristic, named GHOD, is also proposed for benchmarking the performance of GAOD.

A factory hall (102 m $\times$ 24 m) of an automated guided vehicle (AGV) manufacturer and a warehouse (415 m $\times$ 200 m) of a car manufacturer in Belgium are investigated as two case studies, i.e., small-scale and large-scale industrial indoor environment, respectively. Empirically, the feasibility and effectiveness of the OD model and GAOD is validated by measurements in a 10 m $\times$ 10 m empty environment in the factory hall of the AGV manufacturer. Numerically, the effectiveness of GAOD and GHOD is extensively demonstrated in the two investigated environments, without and with the presence of metal racks, in comparison to the random OD solution generation. Compared to GAOD and GHOD, the random OD solution generation outputs up to 60% and 33% more APs, respectively. The superiority of GAOD, compared to GHOD, is demonstrated by the fact that GAOD outputs up to 20% less APs for the same OD problem in a reasonable time span.

The outcome ODGA algorithm can help network managers and plant managers to automatically plan an IWLAN which has high availability under the presence of dominant metal in the environment. Moreover, it can easily be extended to plan other robust wireless networks such as wireless sensor networks.

**Acknowledgement**

This research was supported by the ICON-FORWARD project with imec.